\documentclass[11pt, a4paper, logo, copyright]{eai}
\usepackage{shlab}

\usepackage[authoryear, sort&compress, round]{natbib}
\usepackage[]{mdframed}

\usepackage{dblfloatfix}

\usepackage{amsmath,amsfonts,bm}
\usepackage{multirow}
\usepackage{subcaption}
\usepackage{wrapfig}

\def\eqref#1{equation~\ref{#1}}

\def\1{\bm{1}}

\DeclareMathAlphabet{\mathsfit}{\encodingdefault}{\sfdefault}{m}{sl}
\SetMathAlphabet{\mathsfit}{bold}{\encodingdefault}{\sfdefault}{bx}{n}

\usepackage{listings}
\usepackage{hyperref}
\usepackage{url}
\usepackage{graphicx}
\usepackage{amsmath} %
\usepackage{mathrsfs} %
\usepackage{etoolbox}
\usepackage{cleveref}
\usepackage{tcolorbox}
\usepackage{colortbl}
\usepackage{booktabs}       %
\usepackage{amsfonts}       %
\usepackage{nicefrac}       %
\usepackage{microtype}      %
\usepackage{caption}
\usepackage{xspace} %
\usepackage{subcaption}
\usepackage{algorithm}
\usepackage{algorithmic}
\usepackage{multirow}
\usepackage{lipsum}

\usepackage{booktabs} %
\usepackage{enumitem}%
\setlist[itemize]{noitemsep, topsep=0pt}
\usepackage{enumitem,kantlipsum}

\usepackage{booktabs}
\usepackage{multirow}   
\usepackage{graphicx}    
\usepackage{pifont}
\usepackage{minted}
\usepackage{ulem}
\usepackage{xcolor}
\usepackage{longtable}
\usepackage{booktabs}

\usepackage{listings}
\usepackage{xcolor}
\usepackage[T1]{fontenc}

\lstdefinestyle{yamlstyle}{
    basicstyle=\ttfamily\footnotesize,
    numbers=left,
    numberstyle=\tiny,
    stepnumber=1,
    numbersep=6pt,
    frame=single,
    framerule=0.5pt,
    breaklines=true,
    breakatwhitespace=true,
    tabsize=2,
    captionpos=b,
    keywordstyle=\color{blue},
    commentstyle=\color{gray},
    stringstyle=\color{teal},
}

\newlength\savewidth

\definecolor{baselinecolor}{HTML}{d6eaf8}

\definecolor{mygray}{gray}{0.4}

\AtBeginEnvironment{tcolorbox}{\tiny}

\newcount\Comments  %
\usepackage{color}
\definecolor{darkred}{rgb}{0.9,0,0}
\definecolor{darkgreen}{rgb}{0,0.5,0}
\definecolor{darkblue}{rgb}{0,0,0.7}
\definecolor{purple}{rgb}{.6, 0,.6}
\definecolor{orange}{rgb}{1.0,0.64,0}
\newcommand{\kibitz}[2]{\ifnum\Comments=1\textcolor{#1}{#2}\fi}

\title{InternData-A1:  Pioneering High-Fidelity Synthetic Data \\ for
Pre-training Generalist Policy}

\correspondingauthor{* equal contributions, $\dag$ equal corresponding authors}

\renewcommand{\today}{}

\author[*,1,2]{Yang Tian}
\author[*,1]{Yuyin Yang}
\author[*,1]{Yiman Xie}
\author[*,1]{Zetao Cai}
\author[*,1]{Xu Shi}
\author[1]{Ning Gao}
\author[1]{Hangxu Liu}
\author[1]{Xuekun Jiang}
\author[1]{Zherui Qiu}
\author[1]{Feng Yuan}
\author[1]{Yaping Li}
\author[2]{Ping Wang}
\author[1]{Junhao Cai}
\author[$\dag$,1]{Jia Zeng}
\author[$\dag$,2]{Hao Dong}
\author[$\dag$,1]{Jiangmiao Pang}

\affil[1]{Shanghai AI Laboratory}
\affil[2]{Peking University}

\newcommand{\datasetname}{InternData-A1\xspace}

\begin{document}


\begin{abstract}
Recent works explore how real and synthetic data contribute to Vision-Language-Action (VLA) models' generalization. While current VLA models have shown the strong effectiveness of large-scale real-robot pre-training, synthetic data has not previously demonstrated comparable capability at scale.
This paper provides the first evidence that synthetic data alone can match the performance of the strongest $\pi$-dataset in pre-training a VLA model, revealing the substantial value of large-scale simulation.
The resulting model also exhibits surprisingly zero-shot sim-to-real transfer on several challenging tasks.
Our synthetic dataset, InternData-A1, contains over 630k trajectories and 7,433 hours across 4 embodiments, 18 skills, 70 tasks, and 227 scenes, covering rigid, articulated, deformable, and fluid-object manipulation. It is generated through a highly autonomous, fully decoupled, and compositional simulation pipeline that enables long-horizon skill composition, flexible task assembly,  and heterogeneous embodiments with minimal manual tuning.
Using the same architecture as $\pi_0$, we pre-train a model entirely on InternData-A1 and find that it matches the official $\pi_0$ across 49 simulation tasks, 5 real-world tasks, and 4 long-horizon dexterous tasks.
We release the dataset and will open-source the generation pipeline to broaden access to large-scale robotic data and to lower the barrier to scalable data creation for embodied AI research.

\links{
  \link{data}{Data:InternData-A1}{https://huggingface.co/datasets/InternRobotics/InternData-A1}, 
  \link{homepage}{Homepage}{https://internrobotics.github.io/interndata-a1.github.io/}, 
}

\end{abstract}

\maketitle

\begin{figure}[h]
    \centering
    \includegraphics[width=0.95\linewidth]{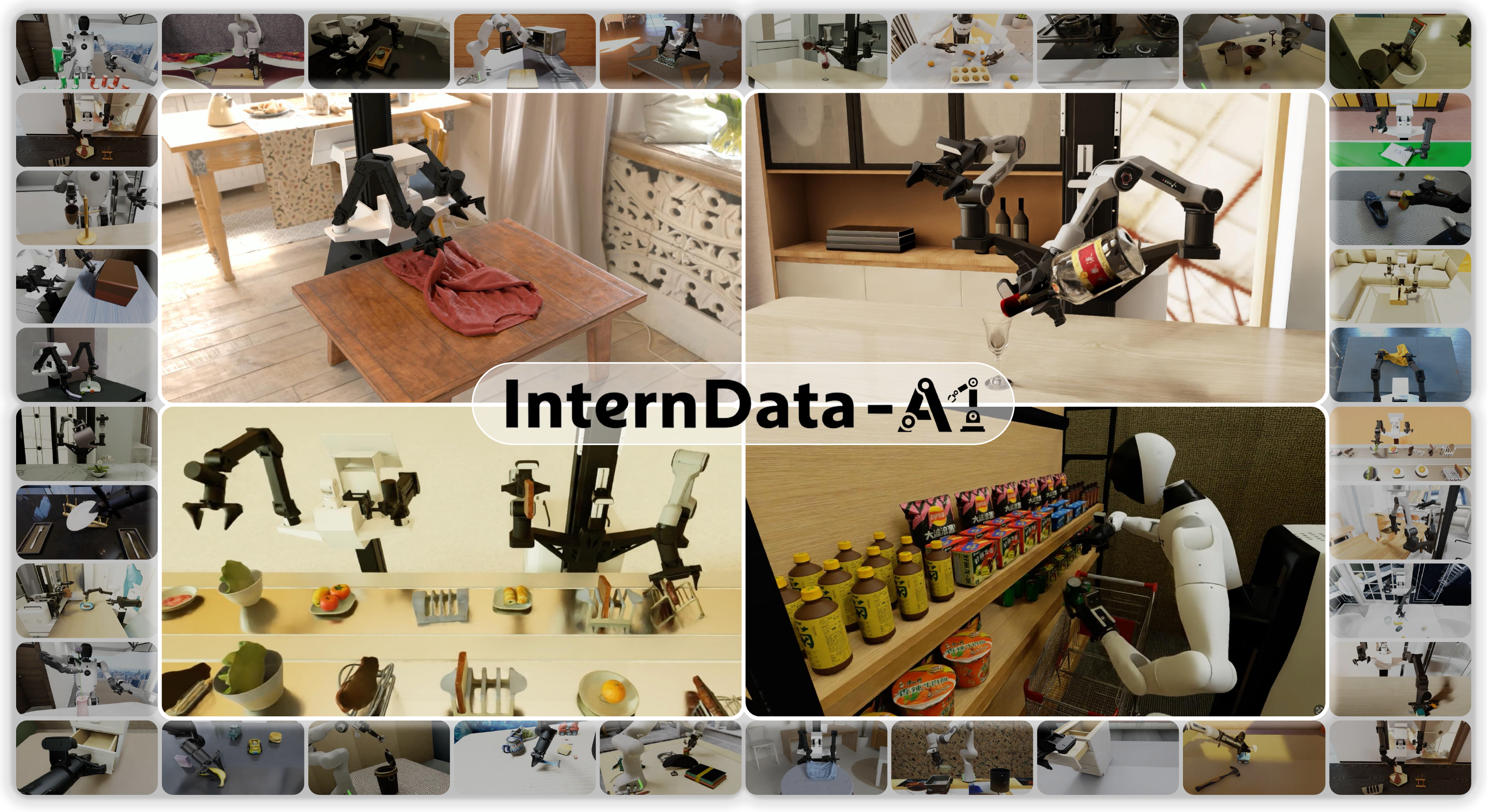}
    \caption{\textbf{InternData-A1} pioneers a large-scale, high-fidelity synthetic dataset with physically faithful, photorealistic rendering, diverse object domains (rigid, articulated, fluid, and deformable), extensive multi-skill tasks, and broad cross-embodiment coverage.} %
    \label{fig:teaser}
\end{figure}


\clearpage
\begin{table}[t!]
\centering
\resizebox{\textwidth}{!}{
\begin{tabular}{l|c|cccc|cc|c|c|c}
\toprule
Dataset & Traj. & Skill & Task  & Scene & Embodiment & Fluid & Deformable & Failure Recovery & Open-Source & Collection \\
\midrule
MimicGen~\citep{mimicgen} & 50k & - & 18 & 1 & 4 & \ding{55}  & \ding{55} & \ding{55}  & \ding{51} & Teleoperation \& Augmentation\\
ManiSkill2~\citep{maniskill2} & 30k & 20 & 20 & - & 1 & \ding{55} & \ding{55} & \ding{55} & \ding{51} &  TAMP \& MPC \& RL
\\
RLBench~\citep{rlbench} & 2.7k & - & 100 & 1 & 1 & \ding{55} & \ding{55} & \ding{55} & \ding{51} & Waypoints \& Motion Planners\\
LIBERO~\citep{libero} & 5k & - & 130 & 20 & 1 & \ding{55} & \ding{55} & \ding{55} & \ding{51} & Teleoperation\\
CALVIN~\citep{calvin} & 37.5k & - & 34 & 4 & 1 & \ding{55} & \ding{55} & \ding{55} & \ding{51} & Teleoperation\\
RoboCasa~\citep{robocasa} & 77k & 8 & 100  & 120 & 2 &  \ding{55} & \ding{55} & \ding{55} & \ding{51} & Teleoperation \& Augmentation \\
GraspVLA~\citep{graspvla} & 10M  & 1 & 1  & 1 & 1 & \ding{55} & \ding{55} & \ding{55} & \ding{55} & Autonomous \\
RoboTwin 2.0~\citep{robotwin2.0} & 100k & - & 50  & 1 & 5 & \ding{55} & \ding{55} & \ding{55} & \ding{51} & Autonomous \\
InternVLA-M1~\citep{internvla-m1} & 244k & 2 & 1  & 1 & 1 & \ding{55} & \ding{55} &\ding{55} & \ding{51} & Autonomous \\
\midrule \midrule
\textbf{\datasetname}  & 630k & 18 & 70 & 227 & 4  & \ding{51} & \ding{51} & \ding{51} & \ding{51} &  Autonomous \\
\bottomrule
\end{tabular}}
\caption{\textbf{Comparison to existing simulation datasets.}}
\label{tab:related_works}
\end{table}

\section{Introduction}
Manipulation lies at the core of embodied intelligence, enabling robots to assist in daily tasks and deliver real economic value. Recent advancements~\citep{pi0,pi0.5,aw,gr00t-n1,cronusvla} demonstrate that large-scale real-robot data can endow Vision-Language-Action (VLA) models with strong generalization across tasks, scenes, and embodiments. However, collecting real data at such scale is resource-intensive: teleoperation demands skilled operators, specialized hardware, and extensive human labor, making large, diverse real-world datasets infeasible for most research groups. 
As a result, the broader community lacks the ability to systematically study the data requirements that underpin successful VLA pre-training.

Simulation offers a promising complementary avenue: its rich asset libraries, controllable scenes, and automated data generation pipelines create the possibility of scaling manipulation data far beyond what is practical with real robots. Yet existing simulated datasets~\citep{mimicgen,robocasa,graspvla,internvla-m1,robotwin2.0} still cover narrow skill sets (primarily pick-and-place), focus mainly on rigid objects, require nontrivial human operation, and rarely validate their effectiveness for large-scale VLA pre-training. This leaves an open question: can high-fidelity synthetic data, when scaled sufficiently in embodiments, scenes, skills, and physical realism, match the pre-training effectiveness of the strongest real-world datasets?

As depicted in~\cref{fig:teaser} and~\cref{tab:related_works}
, we introduce \datasetname, a high-fidelity synthetic manipulation dataset comprising 630k trajectories and 7,433 hours across 4 embodiments, 18 skills, 70 tasks, and 227 scenes, covering rigid, articulated, deformable, and fluid-object interactions.
\datasetname\ is generated through a fully decoupled, autonomous simulation pipeline that separates asset specification, skill policies, task composition, and rendering.
Each task is built by retrieving embodiments, scenes, and objects from an asset library, then composing scripted skill policies that compute and interpolate trajectories conditioned on robot and object states into complete behaviors.
This compositional design flexibly supports bimanual manipulation, multi-robot coordination, and extended multi-stage tasks, while automatically generating object-level annotations, domain randomization, and collision-aware motion plans.
With framework optimizations, the pipeline produces 209.7 hours robot data per day on 8 RTX 4090 GPUs with minimal manual tuning and a cost below 0.003 US Dollars per episode, enabling scalable synthesis of physically and visually faithful demonstrations.
This pipeline underpins the broad task diversity of \datasetname\ and enables systematic VLA validation through large-scale pre-training and sim-to-real evaluation.

We demonstrate that a $\pi_0$ model pre-trained exclusively on \datasetname\ achieves comparable performance to the official $\pi_0$ trained on the closed-source $\pi$-dataset across diverse real-world scenarios.
%
This result, for the first time, establishes that large simulation-only data can match the strongest real-world data for VLA pre-training.
Across 49 simulation, 5 real-world, and 4 long-horizon dexterous tasks, the synthetic-pretrained model closely matches the official $\pi_0$ model.
Furthermore, compared with existing open-source datasets~\citep{aw,oxe,robocasa}, models pre-trained on \datasetname\ consistently outperform counterparts in both simulated and real-world evaluations, highlighting the strength of \datasetname\ for VLA pre-training.

In addition, ten selected simulated tasks achieve direct sim-to-real transfer with an average success rate exceeding 50\%.
Further analysis on four representative tasks reveals that, under well-aligned sim-to-real settings, fewer than 1{,}600 simulated samples can match the performance of 200 real samples, highlighting the high fidelity of \datasetname\ and its minimal sim-to-real gap.

In conclusion, while the $\pi$-dataset remains closed-source, \datasetname\ and its generation pipeline are  open-sourced and reproducible in simulation.
We hope this work offers an accessible supplement to embodied AI research and enables a deeper understanding of data scaling effects in robot learning.

\section{Related Works}
\label{sec:formatting}

\paragraph{Large-scale Robot Datasets.}
Existing large-scale robot datasets fall into two categories: simulation-based~\citep{calvin, rlbench, maniskill2} and real-world datasets~\citep{bridge, galaxea}.
Simulation enables controllable variation in embodiments, scenes, and objects, and supports both teleoperation-based augmentation~\citep{robocasa} and automated pipelines~\citep{graspvla, internvla-m1}. 
However, teleoperation still requires substantial manual effort, and most automated systems mainly target picking tasks. RoboTwin 2.0~\citep{robotwin2.0} recently expanded simulation to bimanual manipulation with cross-embodiment diversity and domain randomization.
Real-world datasets~\citep{oxe, droid, aw, robomind, pi0} offer high-fidelity demonstrations without simulation gaps, but are costly to scale and typically exhibit narrow embodiment coverage, limited scene diversity, or variable data quality. The $\pi$-series datasets~\citep{pi0} demonstrate that large-scale real-robot pre-training can yield strong generalization across tasks, scenes, and embodiments, highlighting the importance of scale and diversity for VLA models.
In contrast to prior simulation datasets,
InternData-A1 provides broad task diversity, heterogeneous embodiments, photorealistic rendering, and long-horizon multi-skill trajectories, enabling systematic study of large-scale synthetic data pre-training for VLA models.

\paragraph{VLA Models.}
The generalization capability of VLA models are primarily determined by the scale, diversity, and source of their pre-training data. Existing approaches fall into three broad categories.
(1) Real-data–only models~\citep{openvla,pi0,RT-1,RT-2,ppi,gr2,gr3} achieve strong in-domain performance~\citep{pi0} but remain tied to specific training environments, often exhibiting limited open-world generalization~\citep{seer,univla}.
(2) Simulation-only models~\citep{graspvla} benefit from massive synthetic trajectories and extensive domain randomization, but typically cover narrow skill families (e.g., grasping), constraining their applicability.
(3) Hybrid models combine simulation and real demonstrations~\citep{internvla-m1,gr00t-n1} or incorporate web-scale VQA data~\citep{pi0.5,EO-1,wall-oss,instructvla} to improve grounding and long-horizon planning.
In this work, we leverage \datasetname\ to provide VLA models with stronger action priors and generalization ability, achieving downstream performance on par with state-of-the-art real data despite the inherent physics and visual gaps of simulation.
Accordingly, we adopt the $\pi_0$ model and its underlying $\pi$-dataset as our primary baselines.

\begin{figure}[t!]
    \centering
    \includegraphics[width=\linewidth]{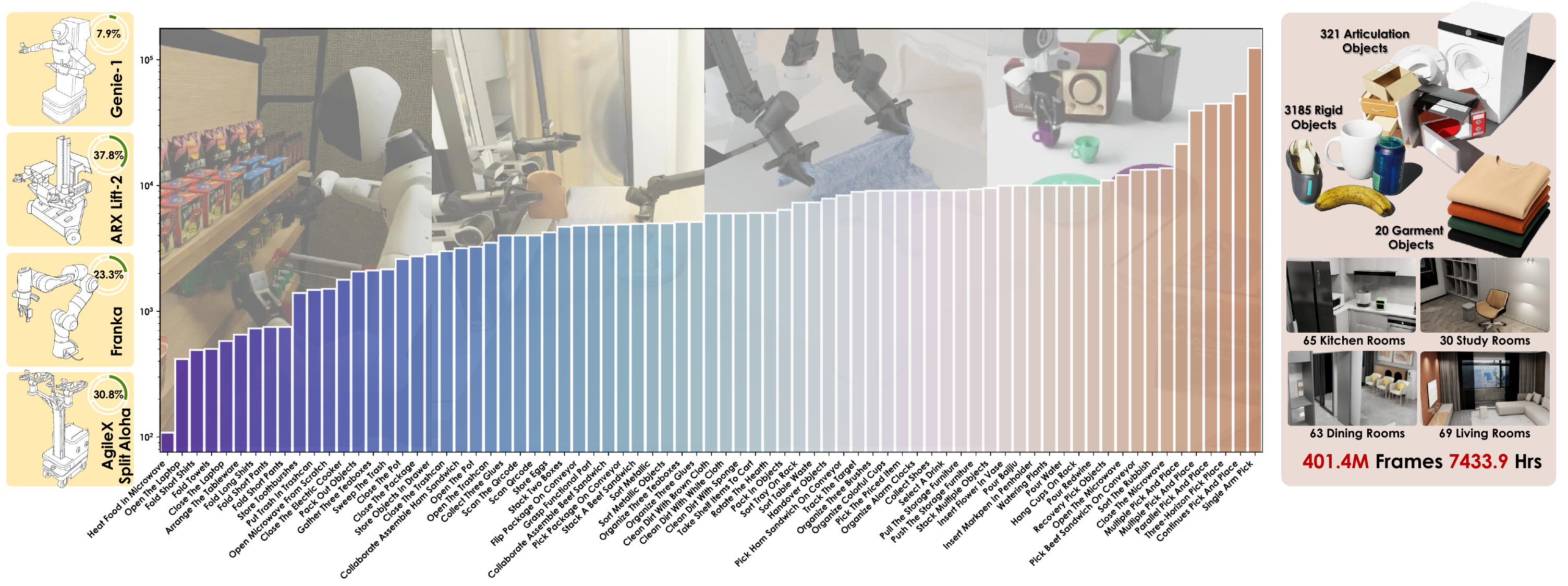}
    \caption{\textbf{Data Statistics.} \datasetname provides 4 single or dual-arm embodiments, 70 diverse tasks, 3185 rigid objects, 321 aritculation objects, 20 garments, and 227 rooms.
    All these elements consist 630k episodes, 401.4M frames and 7433.9 hours.
    }
    \label{fig:data_stats}
\end{figure}
\section{Data Statistics}
In this section, we elaborate on the data statistics from three aspects: embodiments, tasks, and assets, 
with the overall scale and diversity visualized in~\cref{fig:data_stats}.

\noindent \textbf{Embodiments.}
\datasetname\ includes four types of robotic embodiments:  
\textbf{AgiBot Genie-1}, one of the most widely used dual-arm robots in large-scale real robot data collection factories~\citep{aw}, accounts for 7.9\%.  
\textbf{Franka Emika Panda}, one of the most common single-arm manipulators in research laboratories, constitutes 23.3\%.  
We fix Franka on a table to perform tabletop manipulation tasks, including both regular and long-horizon scenarios.  
\textbf{AgileX Split Aloha} equipped with Piper-100 arms, and \textbf{ARX Lift-2} equipped with R5a arms, are two popular embodiments in current VLA~\citep{pi0,pi0.5,xvla} real-world deployments, occupying 30.8\% and 37.8\%, respectively.  
These dual-arm systems enable rich bimanual manipulation, including sequential skills (e.g., make a sandwich) and parallel behaviors (e.g., fold shirts) across diverse scenes.

\noindent \textbf{Tasks.}
To ensure both trajectory-level and task-level diversity,
\datasetname\ is constructed from a comprehensive set of manipulation primitives spanning nearly all fundamental human-like skills—from folding and pouring to rotating and stacking.
By composing these primitives, we design 70 diverse tasks across realistic scenarios, including 4 fluid-related, 4 deformable-object, 15 articulated, and 47 rigid-object tasks.
Notably, unlike prior works~\citep{robotwin2.0,robocasa,rlbench,calvin}, our tasks are not defined by simply varying manipulated objects (e.g., picking hundreds of objects still counts as one task in our statistics);
each task instead specifies a distinct context, composition of atomic skills, and action-space constraint.
\datasetname\ also includes 18 long-horizon tasks, each involving at least three sequential skills, totaling 124,789 trajectories and 141,421,619 frames. 
Beyond diversity, we ensured balance by proportionally mixing data from different tasks, resulting in 56 of the 70 tasks containing between 1,000 and 10,000 trajectories, approaching a near-uniform task distribution.

\noindent \textbf{Assets.}
We provide \textbf{3,185 rigid objects} from 107 categories, sourced from OmniObject3D~\citep{omniobject3d} and Objaverse~\citep{objaverse}.
We aggregate articulated objects from multiple sources, including GRUtopia~\citep{grutopia}, GAPartNet~\citep{gapartnet}, GenSim2~\citep{gensim2}, Infinite Mobility~\citep{im}, and ArtVIP~\citep{artvip}.
In total, this results in \textbf{321 articulated objects} across 14 categories, ranging from small items such as cartons to large household appliances like refrigerators.
For garment assets, we use the EinScan Rigel Pro scanner to digitize \textbf{20 real garments} (short-sleeve shirts, sweaters, shorts, and towels).
%
Regarding the environments, \datasetname\ contains \textbf{227 indoor scenes} (kitchens, study rooms, dining rooms, and living rooms), 
forming a diverse and realistic layout setup.
All these scenes are acquired and segmented from the GRScenes-100 dataset in GRUtopia~\citep{grutopia}.

\begin{figure}[t!]
    \centering
    \includegraphics[width=\linewidth]{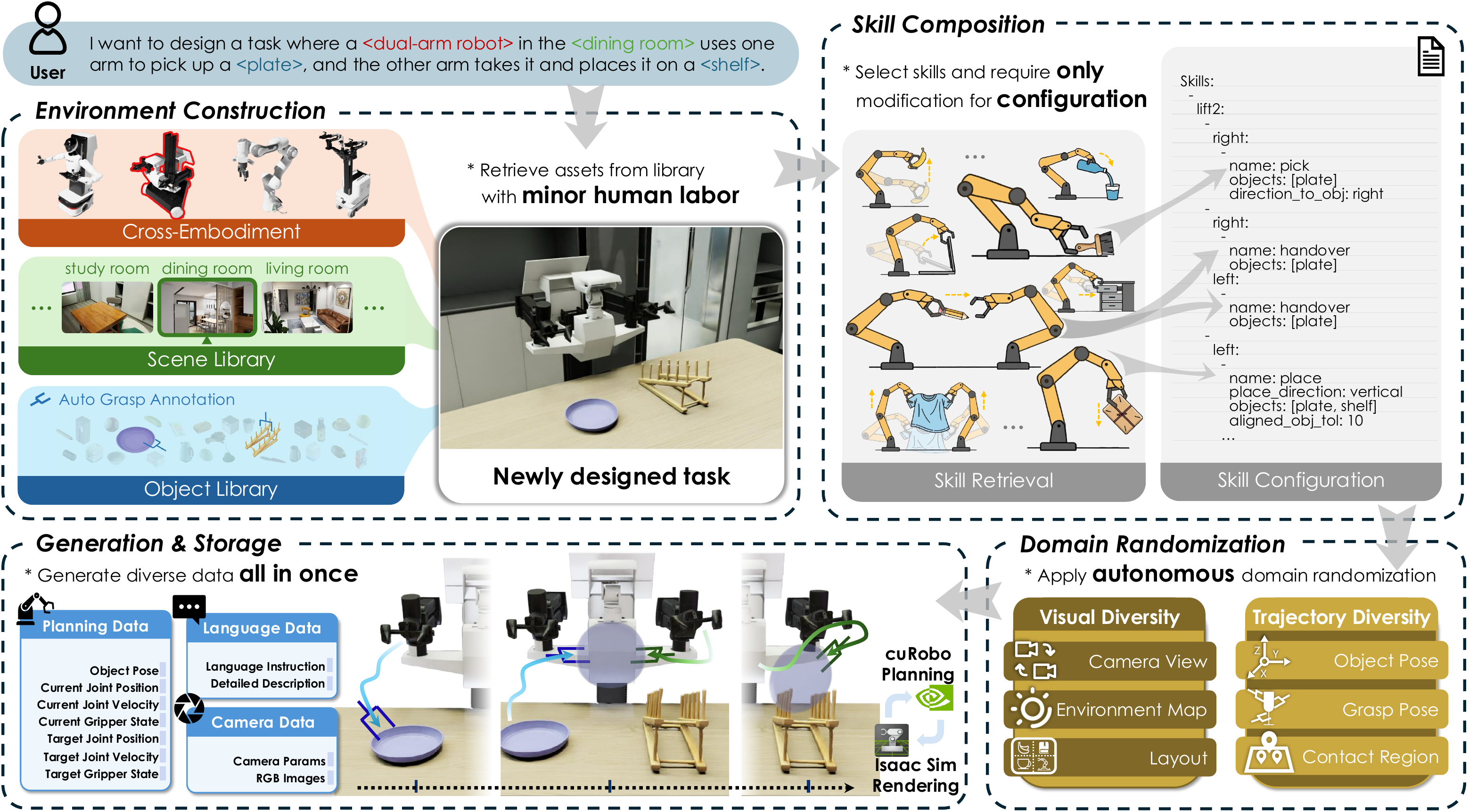}
    \caption{
    \textbf{Data synthesis pipeline of \datasetname.} It consists of four stages: (1) environment construction with selected embodiments, scenes, and objects; (2) task composition using modular atomic skills invoked via simple configuration commands; (3) domain randomization over layouts, object poses, lighting, etc.; and (4) trajectory generation, where CuRobo~\citep{curobo_report23} interpolates dense joint actions, validates them through physics simulation, and renders only successful trajectories into the \texttt{LeRobot} format.}
    \label{fig:data synthesis}
\end{figure}

\section{Data Synthesis}
In this section, we outline the data synthesis pipeline:
we first describe environment construction (robots, scenes, and objects) (Sec~\ref{sec4.1}), then detail how skills constitute complete tasks (Sec~\ref{sec4.2}), and finally explain the domain randomization (Sec.~\ref{sec4.3}), generation, and storage procedures (Sec.~\ref{sec4.4}).

\subsection{Environment Construction}
\label{sec4.1}
As illustrated in~\cref{fig:data synthesis}, given a task description template, we first retrieve task-relevant assets from the library, including robots, scenes, and objects.
The robot asset is a stable USD-type embodiment that has been verified for consistent contact dynamics in robot–object interactions.
The scene asset is a room-level environment, selected and segmented from  GRUtopia~\citep{grutopia}. 
Each scene is annotated with detailed manipulation-area metadata, enabling natural and context-aware task construction.
Our object library spans \textbf{rigid}, \textbf{articulated}, \textbf{deformable}, and \textbf{fluid} categories.
Rigid objects include physical attributes, canonical poses, and automatically generated grasp poses following AnyGrasp~\citep{anygrasp}.
Articulated objects are annotated with accurate joint axes, part poses, and physical parameters (e.g., damping, stiffness) for stable interaction.
Deformable assets are remeshed garments with material and contact properties, simulated using Vertex Block Descent~\citep{VBD} for realistic behavior.
Fluid objects are modeled with particle-based dynamics, where particles are adaptively generated within containers, and liquid surfaces reconstructed via isosurface rendering with diverse PBD materials.
All functional annotations are managed through a unified interface, enabling easy asset configuration with high physical fidelity.

\subsection{Skill Composition}
\label{sec4.2}
Once the environments are prepared, users compose tasks by selecting atomic skills from the skill library.
Each skill is a modular scripted policy that takes object states (poses, joint states), robot states (base and end-effector poses), and user-defined constraints as inputs, and outputs a sequence of waypoints (target end-effector 6D poses).
Waypoints serve as a unified representation, cleanly decoupling high-level skill logic from low-level motion execution.
For example, the \textit{Pick} skill filters grasp candidates for the target object and computes the pre-grasp, grasp, and post-grasp poses, while the articulation-related \textit{Push} skill uses contact annotations to determine pre-contact, contact, and post-contact waypoints.
For tasks with special spatial constraints, we provide a set of constraint options for each skill. 
For instance, in the task \textit{Insert Flower In Vase}, it requires the stem to remain upright.
Users could specify the displacement and rotation between the placed object and container in \textit{Place} skill. 
All these well-engineered skills can be invoked through simple command templates in a configuration file (see~\cref{fig:data synthesis}). 
Since each skill is designed as an automatic state-waypoint mapping, no additional cost is required when varying objects, spatial ranges, scenes or even embodiments.
The only manual effort required is merely the adjustment of spatial ranges. 
Users simply specify which arm (left or right) to use for each skill and organize them sequentially or in parallel for bimanual tasks, and then a long-horizon task unfolds smoothly and systematically.

\subsection{Domain Randomization} 
\label{sec4.3}
To enrich visual diversity, we perturb the primary and wrist camera views within $\pm5^\circ$ rotations and $\pm5$\,cm translations.
We construct a library of 174 environment maps, each with randomized light temperature and intensity to simulate diverse natural illumination conditions.
Target objects can be replaced with others from the same category, while the tabletop and background layouts are also randomized.
To further enhance trajectory diversity, object positions and orientations are sampled within task-specific spatial ranges.
For manipulation functions, we introduce additional randomness to the contact regions.
For instance, our autonomous grasp pose generation pipeline produces millions of grasp candidates; after filtering, the final grasp pose is randomly selected from the top 40 high-confidence candidates following Anygrasp~\citep{anygrasp}.
For articulated and deformable objects, such as microwaves or garments, we expand the contact region into a neighborhood area and sample the contact points randomly.

\subsection{Generation \& Storage} 
\label{sec4.4}
After all preparations, we employ the CuRobo motion planner~\citep{curobo_report23} to interpolate dense joint-space actions between the waypoints generated by each skill.
%
%
For each complete robot episode, we record object metadata, language instructions, multi-view RGB images and camera parameters, as well as robot proprioceptive states and action control labels.
Users can also easily store additional information, such as depth maps, grounding annotations, and bounding boxes, with simple configuration options during recording.
All final data are converted into the standard \texttt{LeRobot} format for VLA pre-training.

\subsection{Framework Optimization}
In traditional synthetic data generation pipelines, trajectory planning and visual rendering are integrated into a single stage. While this architecture is suitable for rapid development and iteration, it exhibits substantial efficiency bottlenecks when scaled to large-scale data generation. The root causes can be summarized as follows:

\begin{enumerate}[leftmargin=*]
    \item \textbf{Declining planning success rate with increasing task complexity.} As task complexity grows, the success rate of trajectory planning decreases significantly. Failed trajectories do not require subsequent visual rendering, yet the single-stage architecture incurs redundant rendering overhead, resulting in unnecessary computational waste.
    \item \textbf{Mismatch in computational characteristics.} Trajectory planning is fundamentally CPU-bound and executed serially, whereas visual rendering relies on GPU-based parallel computation. Executing these heterogeneous workloads in a serial manner leads to poor overall hardware utilization.
\end{enumerate}

To mitigate these bottlenecks, we introduce a multi-level system optimization at the framework level. Our design includes:

\begin{enumerate}[leftmargin=*]
    \item \textbf{Stage decoupling with a pipelined architecture:} Trajectory planning and visual rendering are decoupled into two independent stages, with a pipelined execution mechanism established between the Planner and Renderer.
    \item \textbf{Dynamic resource scheduling:} To address heterogeneous time-cost ratios across different tasks, we incorporate parallel batch processing strategies within both the Planner and Renderer, together with a dynamic scheduling algorithm to maximize resource utilization.
    \item \textbf{Rendering efficiency optimization:} We introduce a stacked rendering (Stack Render) technique to further increase rendering throughput.
    \item \textbf{Cluster stability mechanisms:} To handle stability issues and load imbalance in large-scale cluster deployments, we design a \textit{Balancer} module for load distribution and a \textit{Supervisor} module for monitoring and control, jointly improving cluster utilization and system robustness.
\end{enumerate}

With these optimizations, our pipeline achieves a \textbf{2--3$\times$} end-to-end performance improvement over the baseline. It further supports long-duration stable operation and efficient large-scale synthetic data generation, substantially improving productivity in synthetic data production.

\section{Pre-training Effects}
In this section, we address the central question of whether \datasetname\ can match the real-robot dataset in its capacity to pre-train VLA models.

\begin{figure}[htbp]
    \centering
    \includegraphics[width=\linewidth]{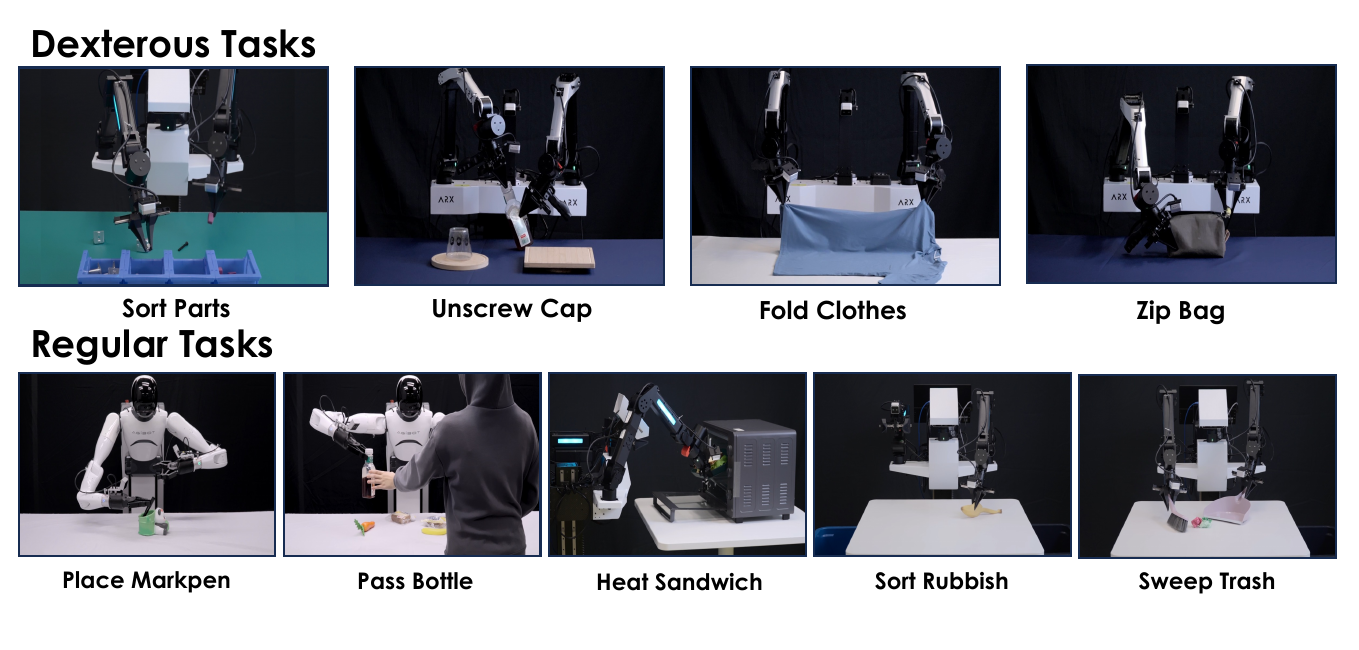}
    \caption{
    \textbf{Real-world setup.}
    We evaluate the model pre-trained on \datasetname\ against the $\pi$-dataset baseline on a suite of nine real-world tasks spanning three robots.
    }
    \label{fig:real_setup}
\end{figure}

\subsection{Comparison with $\pi$-dataset}
\noindent \textbf{Experimental Setup.}
We benchmark against the strongest real-robot data, $\pi$-dataset. 
We adopt the same architecture as $\pi_0$ model,
which consists of a vision-language model (Paligemma~\citep{paligemma}) and a flow-matching-based action expert. 
We pre-train a new $\pi_0$ model from Paligemma weights and a randomly initialized action expert, using solely \datasetname,
and benchmark it against the official $\pi_0$ checkpoint by fine-tuning and evaluating on downstream tasks. 
This comparison reflects the impact of pre-training data quality and provides the fairest available solution, as $\pi$-dataset remains fully closed-source. 
For simulation evaluation, we use 49 bimanual tasks from RoboTwin~2.0~\citep{robotwin2.0} under two difficulty modes: \textit{Easy} (clean) and \textit{Hard} (cluttered). 
We report average success rates over 100 trials across two checkpoints. 
For real-world evaluation, we test on three embodiments: Genie-1, ARX Lift-2 and ARX AC One. 
We design five regular and four dexterous tasks covering articulated manipulation, garment folding, contact-rich control, precision grasping, and long-horizon sequences. 
Each task is evaluated over 30 trials.
In the following results, we use \textit{$\pi_0$ (Scratch)} to denote the model without being pre-trained, \textit{$\pi_0$} for the official checkpoint, and \textit{$\pi_0$ (\datasetname)} for the model trained exclusively on our dataset.\\

\begin{table}[htbp]
\centering
\setlength{\tabcolsep}{5pt}
\scriptsize
\resizebox{\textwidth}{!}{
\begin{tabular}{l|cc|cc|cc|cc|cc|cc|cc}
\toprule
\multirow{2}{*}{Method} &
\multicolumn{2}{c|}{Hanging Mug} &
\multicolumn{2}{c|}{Lift Pot} &
\multicolumn{2}{c|}{Pick Dual Bottles} &
\multicolumn{2}{c|}{Place Object Stand} &
\multicolumn{2}{c|}{Shake Bottle Horizontally} &
\multicolumn{2}{c|}{Turn Switch} &
\multicolumn{2}{c}{Avg. (49 Tasks)} \\
\cmidrule(lr){2-15}
 & Easy & Hard & Easy & Hard & Easy & Hard & Easy & Hard & \ \ \ \ Easy & \ \ \ \ Hard & Easy & Hard & Easy & Hard \\
\midrule
$\pi_0$ (Scratch) & 5.0\%  & 2.0\% & 26.5\% & 0.0\% & 1.5\% & 0.5\% & 9.0\% & 0.0\% & \ \ \ \ 55.0\% & \ \ \ \ 2.0\% & 9.0\% & 9.5\% & 23.5\% & 2.5\% \\
$\pi_0$ & 11.0\%  &  6.5\% & 17.0\% & 1.5\% & 58.0\% & 16.0\% & 43.0\% & 14.0\% & \ \ \ \ 96.5\% & \ \ \ \ 55.0\% & 27.5\% &  30.0\% & 55.0\% & 20.0\% \\
\textbf{$\pi_0$ (\datasetname)} & \textbf{24.5\%} & \textbf{20.0\%} & \textbf{63.5\%} & 
\textbf{2.5\%} & 
\textbf{62.0\%} & 
\textbf{19.0\%} & 
\textbf{48.5\%} & \textbf{29.5\%} & \ \ \ \ \textbf{98.0\%} & \ \ \ \ \textbf{64.0\%} & \textbf{40.5\%} & \textbf{32.5\%} & \textbf{60.0\%} & \textbf{26.5\%} \\
\bottomrule
\end{tabular}
}
\normalsize
\label{tab:compared_with_pi_sim}
\caption{\textbf{Comparison to $\pi$-dataset in simulations.} $\pi_0$ model pre-trained on \datasetname outperforms the official $\pi_0$ model by 5\% and 6.5\% across 49 tasks in easy and hard settings respectively, showing the effectiveness of \datasetname. }
\end{table}

\textbf{Evaluating in Simulation.} ~\cref{tab:compared_with_pi_sim} summarizes the results over 49 tasks and 19,600 rollouts. 
Under an identical post-training process, models pre-trained on \datasetname\ consistently outperform those trained on the $\pi$-dataset, achieving 6\% higher success rates in the \textit{Easy} mode and 5\% higher in the \textit{Hard} mode. 
The improvement in hard settings (clean expert demonstrations and domain-randomized evaluation) indicates that the robustness to visual and spatial variations gained from InternData-A1’s extensive domain randomization persists even when downstream fine-tuning uses only clean, non-randomized data. 
Moreover, since \datasetname\ consists of diverse tasks freely composed from repeated atomic skills across varied contexts, we believe it provides a transferable and broader action prior for downstream tasks sharing similar skill components.
This is evidenced by the six representative tasks in~\cref{tab:compared_with_pi_sim}, which demand robust combinations of pick, place, move, handover, and articulation-related skills.
Overall, \datasetname\ improves over \textit{Paligemma} by 36.5\% in \textit{Easy} and 24.0\% in \textit{Hard} mode, demonstrating the strong pre-training effectiveness of \datasetname.
\\

\begin{figure}[htbp]
  \centering
  \includegraphics[width=\linewidth]{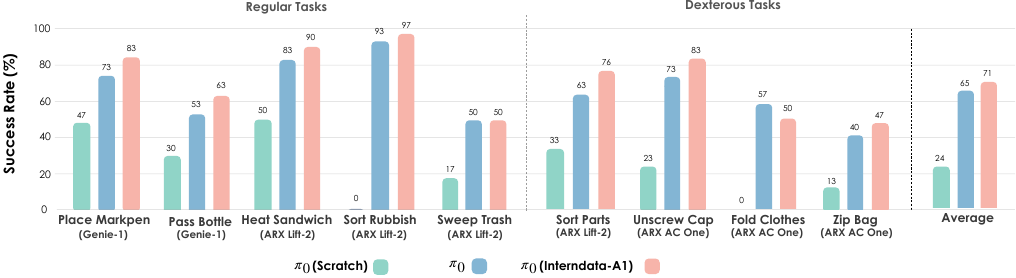} 
  \caption{
    \textbf{Comparison to $\pi$-dataset in real-world tasks.}
    \datasetname\ achieves performance comparable to $\pi$-dataset across 9 real-world tasks, including 4 dexterous ones, demonstrating its strong pre-training capability.
  }
  \label{fig:compared_with_pi_real}
\end{figure}

\textbf{Evaluating over Regular Tasks.} We first compare \datasetname\ and $\pi$-dataset on five representative real-world base tasks across two embodiments, covering articulation (\textit{Heat Sandwich}), basic pick-and-place (\textit{Sort Rubbish}, \textit{Place Markpen}), human–robot interaction (\textit{Pass Bottle}), and tool use (\textit{Sweep Trash}). 
Across both in-distribution and out-of-distribution tasks, \datasetname\ consistently outperforms $\pi$-dataset by a 6.2\% margin. 
In particular, consistent with superior performance on RoboTwin, the rich  basic skill trajectories in \datasetname\ (e.g., pick, place, and move) promote performance on tasks composed mainly of these skills (e.g., \textit{Place Markpen} and \textit{Pass Bottle}).
These results send an encouraging message: although simulations inevitably differ from real-world dynamics, many manipulation tasks exhibit natural robustness, as robotic hardware tolerates minor control inaccuracies, VLA models learn abstract visuomotor representations, and many tasks allow for approximate contact strategies. Consequently, large-scale simulated data can still provide transferable inductive priors for real-world manipulation.
\\

\textbf{Evaluating over Dexterous Tasks.}
We further evaluate the pretrained policy on four long-horizon dexterous tasks—folding garments, sorting industrial parts, unscrewing bottle caps, and zipping bag, directly challenging the core strengths of the official $\pi_0$ model. 
To ensure fair evaluation, we introduce a new embodiment, ARX AC One, which is unseen in both \datasetname\ and $\pi$-dataset. 
Despite the presence of novel objects, skills, and embodiments, \datasetname\ achieves performance comparable to $\pi$-dataset, as shown in the right panel of~\cref{fig:compared_with_pi_real}. 
We suppose that, beyond the comprehensive coverage of atomic skills in \datasetname\ as discussed earlier, the substantial proportion of long-horizon tasks further enables the VLA model to explore a broader action space and develop continuous control abilities, which can then be effectively transferred to novel tasks and embodiments.

\begin{table}[htbp]
\centering
\begin{tabular}{l|c|cc|cc}
\toprule
\multirow{2}{*}{Dataset} &
\multirow{2}{*}{Domain} &
\multicolumn{2}{c|}{49 Sim Tasks} &
\multicolumn{2}{c}{2 Real Tasks} \\
\cmidrule(lr){3-4} \cmidrule(lr){5-6}
 & & Easy & Hard  & Sort Rubbish & Pass Bottle \\
\midrule
OXE~\citep{oxe} & Real & 32.5\% & 11.0\% & 40.0\% & 36.7\% \\
Agibot World~\citep{aw} & Real & 52.5\% & 12.0\% & 53.3\% & 56.7\% \\
RoboCasa~\citep{robocasa} & Sim & 50.0\% & 11.0\% & 23.3\% & 13.3\% \\
\textbf{\datasetname} & \textbf{Sim} & \textbf{60.0\%} & \textbf{26.5\%} & \textbf{90.0\%} & \textbf{60.0\%} \\
\bottomrule
\end{tabular}
\label{tab:compared_with_opensource}
\caption{\textbf{Performance comparison of models pretrained on \datasetname\ against other open-source datasets.}}
\end{table}

\subsection{Benchmarking against open-source datasets}
We further compare \datasetname\ with previous open-source datasets, including two real-world datasets OXE~\citep{oxe} and Agibot World~\citep{aw}, and one simulation dataset, RoboCasa~\citep{robocasa}.
Due to resource constraints, 
we pre-train the $\pi_0$ model exclusively on each dataset for 500k iterations and evaluate performance on 49 simulated tasks and 2 real-world tasks.
As shown in~\cref{tab:compared_with_opensource}, \datasetname\ demonstrates clear advantages across both simulation and real domains.
In regard to RoboCasa~\citep{robocasa},
while the RoboCasa demonstrates competitive performance in simulation evaluations, trailing InternData-A1 by only 10\%, its performance significantly drops in real-robot evaluation. In these real-world evaluations, InternData-A1 achieves an average improvement of 57.7\% over RoboCasa.
We attribute this significant gain to \datasetname’s highly photorealistic rendering and abundant data amounts.
Together with the comparison against the closed-source $\pi$-dataset, these results further validate the effectiveness of \datasetname\ as a strong pre-training source for VLA models.

\section{Data Analysis}
In this section, we aim to address two widely discussed questions regarding large-scale simulation data:  
(1) How well do the tasks in \datasetname\ transfer from simulation to the real world?  
(2) Which components of \datasetname\ contribute most to effective pre-training?

\begin{figure}[htbp]
    \centering
    \includegraphics[width=\linewidth]{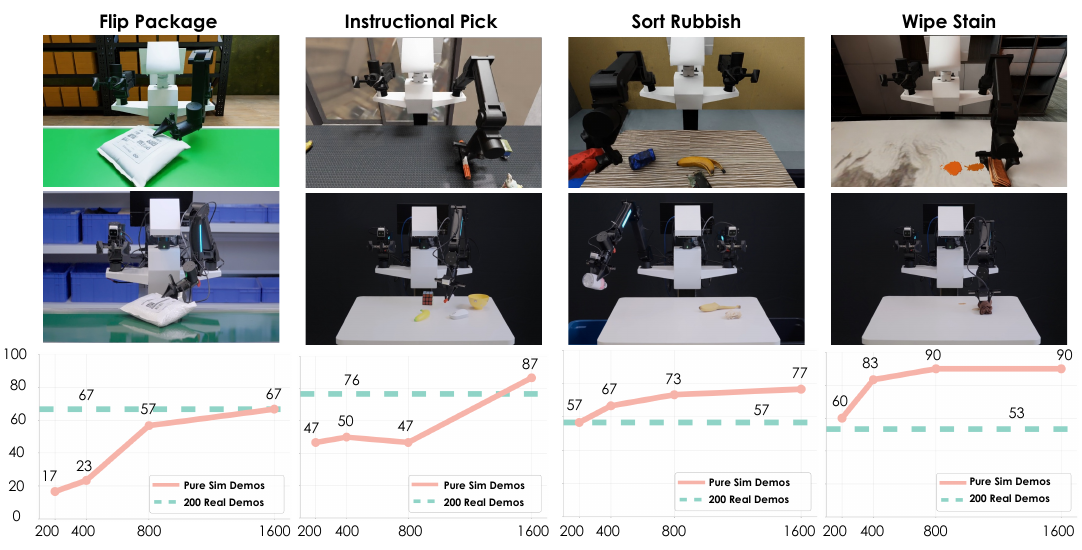}
    \caption{\textbf{Sim-to-real experimental setup.}
    Four tasks are evaluated via direct sim-to-real transfer, and simultaneously, the performance of training with simulation data is compared with that of training with real-world data.}
    \label{fig:sim2real_setup}
\end{figure}

\textbf{Pure Simulation versus Pure Real.}
We investigate the zero-shot sim-to-real transfer ability on four representative tasks in \datasetname. Starting from the same $\pi_0$ (\datasetname) checkpoint, we post-train on 200–1{,}600 simulated episodes and 200 real episodes for each task, and evaluate each resulting policy over 30 rollouts.
As shown in~\cref{fig:sim2real_setup}, for regular tasks such as \textit{Sort Rubbish} and \textit{Wipe Stain}, which mainly involve basic skills (pick, place, move), 200 simulated episodes already achieve performance comparable to 200 real ones.
For more complex tasks such as \textit{Flip Package} and \textit{Instructional Pick}, which require dynamic-object manipulation and language grounding, around 1{,}600 simulated episodes are needed to match real-data performance.
In other words, the simulation-to-real data ratio for equivalent performance narrows to within 8:1—and in some cases, even approaches 1:1.
Although complex tasks demand more simulated data, our efficient synthesis pipeline achieves this at far lower time and economic costs than real-world collection. 
Moreover, we find that strong zero-shot sim-to-real performance does not require exact replicas of backgrounds, lighting, object textures, or table layouts. As shown in~\cref{fig:sim2real_setup}, coarse alignment is sufficient, as long as camera views and joint action spaces are closely aligned. 
This robustness arises from \datasetname’s photo-realistic rendering and extensive domain randomization, which narrow the visual gap and enable the policy to ignore irrelevant discrepancies between simulation and real environments.

\begin{figure}[htbp]
  \centering
  \includegraphics[width=\linewidth]{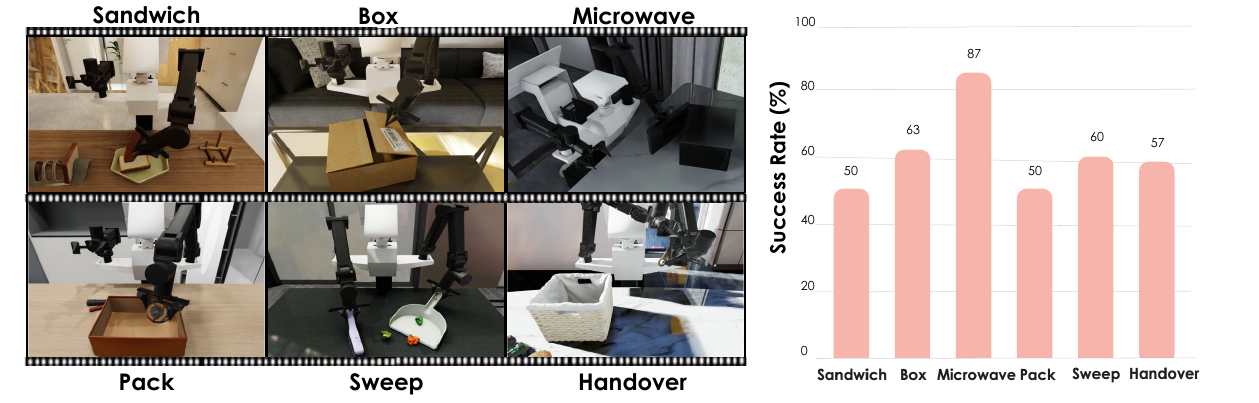}  
  \caption{\textbf{Additional sim-to-real performance.}
Six tasks involving repeated pick-place, articulation, and bimanual coordination achieve over 50\% success rates with only 500 simulated episodes.}
  \label{fig:more_sim2real_tasks}
\end{figure}

\noindent \textbf{Additional Sim-to-Real Tasks.}
To further examine the real-world potential of \datasetname, we evaluate six additional tasks using 500 simulated episodes post-trained on $\pi_0$ (\datasetname) and tested over 30 rollouts.
As shown in~\cref{fig:more_sim2real_tasks}, the model achieves 50\% success rates in pick-and-place tasks (\textit{Make Sandwich}, \textit{Pack}), 63\% and 87\% in articulation operations (\textit{Close Box}, \textit{Close Microwave}), and 60\% and 57\% in bimanual coordination tasks (\textit{Sweep}, \textit{Handover}).
Together with the four tasks in~\cref{fig:sim2real_setup}, ten representative tasks out of seventy achieve high sim-to-real success without any real data — a strong reflection of the data quality and minimal visual–physical gap in \datasetname.
Moreover, \datasetname is the first to demonstrate that, beyond single pick tasks as in GraspVLA~\citep{graspvla}, diverse and complex tasks can also be directly transferred via VLA from simulation to reality.
These include multi-skill operations, articulation, bimanual manipulation, and even \textit{Flip Package} in logistics scenes, which was previously realized using purely real data by the Helix~\citep{openhelix}  from the Figure Company.
These findings highlight the strong potential of large-scale simulation data for VLA deployment in real world.

\subsection{Data Component Ablation}

\begin{table}[htbp]
\centering
\setlength{\tabcolsep}{3pt}
\scriptsize
\resizebox{0.9\linewidth}{!}{
\begin{tabular}{c|c|c|c|c}
\toprule
Full & w.o. PnP. 
& w.o. Art. & w.o. Base. & w.o. Long.   \\
\midrule
Easy / Hard & Easy / Hard & Easy / Hard &  Easy / Hard &  Easy / Hard \\
\midrule
 \textbf{58.0\% / 25.0\%} & 57.0\% / 22.5\% & 55.5\% / 19.5\% & 52.5\% / 20.5\% & 54.0\% / 19.0\%  \\
\bottomrule
\end{tabular}
}
\normalsize
\label{tab:dataset_ablation}
\caption{
\textbf{Data component ablation results.} ``Pnp'', ``Art'', ``Base'', and ``Long'' represent pick-and-place tasks,  articulation manipulation tasks, base tasks, and long-horizon task, respectively.}
\end{table}

\textbf{What Matters in \datasetname's Composition.}
To uncover each component's contribution to pre-training, we divide \datasetname\ into four parts: large-scale pick-and-place tasks (PnP, 30.61\%), large-scale articulation manipulation (Art, 11.67\%), base tasks involving fewer than three skills beyond simple pick-and-place (Base, 35.95\%), and long-horizon tasks containing at least three skills (Long, 21.77\%).
Due to resource constraints, we pre-train $\pi_0$ for 0.5 epoch using both the full dataset and ablated versions (each with one component removed), and evaluate them on RoboTwin 2.0.
As shown in~\cref{tab:dataset_ablation}, the full-data model achieves the highest success rates, while removing any component causes a clear performance drop, indicating all four components contribute meaningfully.

Two observations stand out.
First, although PnP (30.61\%) and Base (35.95\%) dominate the dataset, removing Base or Long tasks leads to a more significant decline than removing PnP.
This suggests that for VLA pre-training, datasets composed solely of pick-and-place tasks~\citep{graspvla, internvla-m1} are insufficient; maintaining task diversity, i.e., multi-skill compositions, is more beneficial for VLA training.
Second, excluding articulation tasks, despite their smaller scale and limited object diversity, also yields a larger drop than excluding PnP.
We suspect that articulated manipulation reaches a more diverse action space, as they are often influenced by diverse joint geometries and distorted arm configurations (e.g., pushing a distant drawer).
At a higher level, combining these two findings, we hypothesize that \textbf{trajectory diversity} may serve as the core drive of effective pre-training.
We leave a rigorous investigation for future research.

\section{Conclusion and Limitation}
This paper introduces InternData-A1, a large-scale, high-fidelity synthetic dataset for robotic manipulation. 
Using this dataset, we demonstrate for the first time that pre-training a Vision-Language-Action (VLA) model exclusively on synthetic data can match the performance of the leading VLA model, revealing the substantial potential of large-scale simulation.
By open-sourcing the dataset and its generation pipeline, we aim to lower the barrier to large-scale robotic data access for the embodied AI community.

\noindent\textbf{Limitation.} Due to the limitations of physics simulators, it is challenging to simulate highly dexterous tasks, such as tying shoelaces or threading a needle. 
Future work will expand the dataset's task diversity and dexterity, aiming to further establish large-scale simulation data as a cornerstone for the advancement of VLA models.

\section{Acknowledgement}
We acknowledge Hui Wang, Xing Shen, Baole Fang, and Yunsong Zhou for their efforts in the simulation and synthesis of deformable object manipulation. 
We thank Zeyu He, Yuanzhen Zhou, and Hengjie Li for their contributions in optimization and acceleration of data synthesis pipeline. 
We are grateful to Jiafei Cao for supporting the real-world experimental infrastructure, and to Liang Pan, Runyi Yu, Yuxi Wei, and Jingbo Wang for their pioneering work on early-stage development of the data synthesis pipeline.
Finally, we acknowledge Linning Xu for her exceptional work in drafting and refining the manuscript.

\clearpage
\bibliography{refs}
\appendix
\newpage

\begin{longtable}{p{3.8cm} c c c c c}
\caption{\textbf{Task Statistics Across Robots.}} \\
\toprule
Task Name  & Franka & ARX Lift-2 & Agilex Split Aloha & Genie-1 & Sum  \\
\midrule
\endfirsthead

\toprule
Task Name  & Franka & ARX Lift-2 & Agilex Split Aloha & Genie-1 & Sum \\
\midrule
\endhead

\midrule
\multicolumn{6}{r}{Continued on next page} \\
\endfoot

\bottomrule
\endlastfoot

\multicolumn{5}{l}{\textbf{Articulation Tasks (11.67\%)}} & \textbf{74{,}415} \\
\midrule
Close The Electric Cooker & 1776 &  & & &  \\
Close The Laptop & 578 &  & & & \\
Close The Pot & 2595 &  & & & \\
Close The Trashcan & 2996 &  & & & \\
Close The Microwave & 2496 & 6831 & 4367 & & \\
Open The Laptop & 416 &  & & & \\
Open The Pot & 3250 &  & & & \\
Open The Trashcan & 3507 &  & & & \\
Open The Microwave & 2139 & 5148 & 4817 & & \\
Pull The Storage Furniture & & 4368 & 4776 & & \\
Push The Storage Furniture & & 4548 & 4775 & & \\
Rotate The Hearth & & 6046 & 4653 & & \\
Open Microwave From Scratch & & & 1501 & & \\
Heat Food In Microwave & & 108 & & & \\
Close The Package & & 2724 & & & \\

\midrule
\multicolumn{5}{l}{\textbf{Long-horizon Tasks (21.77\%)}} & \textbf{138{,}782} \\
\midrule
Clean Dirt With Brown Cloth & & 3000 & 3000 & & \\
Clean Dirt With Sponge & & 3000 & 3000 & & \\
Clean Dirt With White Cloth & & 3000 & 3000 & & \\
Collect Three Glues & & 2000 & 2000 & & \\
Gather Three Teaboxes & & 110 & 2000 & & \\
Handover Objects & & 7863 & & & \\
Pack In Objects & & 4052 & 2345 & & \\
Pack Out Objects & & & 2070 & & \\
Sort The Rubbish & & 4579 & 8860 & & \\
Stack Multiple Objects & & 4867 & 4664 & & \\
Sweep The Trash & 2344 & 528 & 1626 & & \\
Put Trash In Trashcan & 1480 & & & & \\
Collaborate Assemble Beef Sandwich & & 4854 & & & \\
Stack A Beef Sandwich & & 4271 & 670 & & \\
Store Objects In Drawer & & 2822 & & & \\
Collaborate Assemble Ham Sandwich & & 3168 & & & \\
Continues Pick And Place & 20036 & 15000 & 18573 & & \\

\midrule
\multicolumn{5}{l}{\textbf{Base Tasks (35.95\%)}} & \textbf{229{,}168} \\
\midrule
Track The Target & 2959 & 2954 & 3000 & & \\
Organize Three Brushes & 5064 & 2000 & 2000 & & \\
Organize Alarm Clocks & 5111 & 2000 & 2000 & & \\
Organize Colorful Cups & 5097 & 2000 & 2000 & & \\
Organize Three Glues & 5120 & & & & \\
Collect Shoes & 5114 & 2000 & 2000 & & \\
Organize Three Teaboxes & 5119 & & & & \\
Sort Table Waste & 5117 & 214 & 2000 & & \\
Store Eggs & & & & 4244 & \\
Take Shelf Items To Cart & & & & 6040 & \\
Pick Beef Sandwich On Conveyor & & 6658 & 6647 & & \\
Pick Ham Sandwich On Conveyor & & 4092 & 4220 & & \\
Fold Long Shirts & & 731 & & & \\
Fold Short Shirts & & 492 & & & \\
Fold Towels & & 500 & & & \\
Fold Short Pants & & 750 & & & \\
Flip Package On Conveyor & & 4806 & & & \\
Pick Package On Conveyor & & 4900 & & & \\
Hang Cups On Rack & & 5000 & 5000 & & \\
Insert Flower In Vase & & 5000 & 4986 & & \\
Insert Markpen In Penholder & & 5000 & 5000 & & \\
Pour Baijiu & & 4999 & 4999 & & \\
Pour Redwine & & 5000 & 5000 & & \\
Pour Water & & 5000 & 5000 & & \\
Pick The Priced Item & 5105 & 2000 & 2000 & & \\
Select A Drink & 5121 & 2000 & 2000 & & \\
Stack Two Boxes & & 2270 & 2429 & & \\
Sort Tray On Rack & & 3851 & 3444 & & \\
Store Toothbrushes & & 1396 & & & \\
Arrange The Tableware & & 650 & & & \\
Recovery Pick Objects & & 10969 & & & \\
Watering Plants & & 5000 & 5000 & & \\
Scan The QRcode & & & & 4000 & \\
Sort Metallic Objects & & 2500 & 2500 & & \\
\midrule
\multicolumn{5}{l}{\textbf{Pick and Place Tasks (30.61\%)}} & \textbf{195{,}133}\\
\midrule
Single Arm Pick & 24598 & 38865 & 39219 & 21695 &  \\
Parallel Pick And Place & & 15687 & 18497 & 10381 & \\
Grasp Functional Part & & & & 4833 & \\
Multiple Pick And Place & 21358 & & & & \\
\midrule
\multicolumn{5}{l}{\textbf{Overall Trajectories}} & \textbf{637{,}498} \\
\midrule
\multicolumn{5}{l}{\textbf{Overall Frames}} & \textbf{401{,}430{,}981} \\
\midrule
\multicolumn{5}{l}{\textbf{Overall Hours}} & \textbf{7433.91}
\label{tab:supp:table statistics}
\end{longtable}

\section{Detailed Data Statistics}
We report the complete dataset statistics in~\cref{tab:supp:table statistics}. 
In total, the dataset contains \textbf{4} embodiments, \textbf{70} tasks, \textbf{637,498} trajectories, \textbf{401,430,981} frames, and \textbf{7,433.91} hours of interaction data.
As outlined above, the dataset is organized into four categories: \textbf{Articulation}, \textbf{Long-horizon}, \textbf{Base}, and \textbf{Pick and Place}. 
These categories comprise \textbf{74,415}, \textbf{138,782}, \textbf{229,168}, and \textbf{195,133} trajectories, accounting for \textbf{11.67\%}, \textbf{21.77\%}, \textbf{35.95\%}, and \textbf{30.61\%} of the dataset, respectively.
For each task, we report the exact number of trajectories contributed by each embodiment. 
See~\cref{tab:supp:table statistics} for detailed per-task and per-embodiment statistics.

\section{Detailed Data Synthesis}
We present an example task config below.
Following the exact task config, we elaborate on each part in our data synthesis in detail.

\subsection{Environment Construction}
As shown in the configuration, we set the room environment using \texttt{defaults/arenas@arena} and select a dining-room layout in the outer script.
We load the Agilex Split Aloha robot—one of our four embodiments—and specify its motion planner via the \texttt{robot\_file}.
We then retrieve two task-relevant assets, the plate and plate shelf, from our asset library, where each asset is automatically annotated with gravity parameters, collision properties, and grasp poses.
After obtaining the two objects, we assign their initial translations and orientations. 
Objects belonging to the same category share a unified canonical pose definition.

\subsection{Skill Composition}
As shown in the configuration—particularly in the \texttt{skills/split\_aloha} section—we construct tasks by composing skills either sequentially or in parallel. 
Users can simply copy and paste different skill blocks to assemble a task. 
For example, a complete task may be formed by chaining together \texttt{pick}, \texttt{goto\_pose}, \texttt{pick}, \texttt{gripper\_action} (close or open), \texttt{home}, and \texttt{place}. 
The framework supports both sequential execution and parallel execution (e.g., one gripper opens while the other closes), enabling users to specify diverse task requirements. 
Users may also define task-level constraints; for instance, in a placement operation, we enforce \texttt{align\_pick\_obj\_axis} and \texttt{align\_place\_obj\_axis} to be parallel to ensure accurate insertion.
Similarly, \texttt{x\_ratio\_range} and \texttt{y\_ratio\_range} can be used to specify the target insertion layer.
All script-level policies have undergone substantial refinement. 
For manual tuning, users may configure grasp-pose filtering rules (\texttt{filter\_x\_dir}, \texttt{filter\_y\_dir}, \texttt{filter\_z\_dir}) and adjust parameters such as \texttt{post\_grasp\_offset\_min}, and \texttt{place\_z\_offset} to ensure stable grasping and placement while avoiding unsafe motions.

\subsection{Domain Randomization}
For visual domain randomization, we provide options in \texttt{env\_map}, allowing light intensity and rotation to be perturbed within predefined ranges.
We also support camera extrinsic randomization, where camera poses are perturbed by up to $5^\circ$ in rotation and 5\,cm in translation. 
Room scenes can be randomized by sampling from the specified room types. 
For objects, replacements can be sampled from assets within the same category.
At the trajectory level, we define a spatial region in which target objects and robots are initialized with randomized poses for each episode. 
Additionally, the \texttt{robots} configuration allows specifying the mean and standard deviation of the home configuration, enabling diverse initial joint states.
Within the skill definitions, we further introduce loose filtering ranges for grasp and placement poses. 
All poses that satisfy these constraints are retained, and a final pose is selected randomly. 
Together, these mechanisms significantly enhance trajectory diversity within each task.

\section{Policy Training Details}

During real-world training, we pretrain a new $\pi_0$ model, initialized with Paligemma weights and a scratched action expert, on \datasetname for 680k iterations using 32 A100 GPUs (closely matches the 700k iteration steps of the official $\pi_0$ checkpoint trained on the $\pi$-dataset). For 10 sim-to-real experiments and 9 real-world tasks, we start from the 680k $\pi_0$(\datasetname) checkpoint and perform post-training for 30k iterations on 8 GPUs for regular tasks and sim-to-real tasks. For dexterous tasks, we trained for 100k iterations. Key training hyperparameters are summarized in ~\cref{tab:hyperparameters}.

\begin{table}[htbp]
\centering
\begin{tabular}{l c c}
\toprule
\textbf{Hyperparameters} & \textbf{Pre-training} & \textbf{Fine-tuning} \\
\midrule \midrule
Batch Size(Total) & 512 & 128 \\
\midrule
Learning Rate & 5e-5 & 2.5e-5 \\
\midrule
Learning Rate Schedule & Constant & Cosine Decay \\
\midrule
Training Steps & 680k & 30k(Regular)/100k(Dexterous) \\
\bottomrule
\end{tabular}
\caption{\textbf{Training hyperparameters.}}
\label{tab:hyperparameters}
\end{table}

\section{Real-World Experiments}
In this section, we describe the real-world and sim-to-real tasks in detail.
For both experiments, we post-train a JAX-version $\pi_0$ (\datasetname) model for 30k iterations and use the 30k checkpoint for evaluation.
For each task, we define 15 evaluation settings, and to reduce stochasticity, we run two trials per setting. 
In total, each task is evaluated with 30 rollouts, and we report the average success rate.

\subsection{Real Task Description}
\textbf{Place Markpen.} 
The Genie-1 robot is required to pick a black marker with its right arm and place it into a pen holder.
This task evaluates the model's fundamental pick-and-place capabilities. 
A trial is considered successful only if the marker is placed precisely and fully inside the pen holder.
\\
\textbf{Pass Bottle.}
The Genie-1 robot is required to pick up a black tea bottle, lift it upright, and hand it to a nearby person with the right arm. 
The robot may release its gripper only when the human presents their hand.
This task evaluates the model's fundamental abilities in picking, lifting, and human–robot interaction. 
A trial is considered successful only if the bottle is successfully transferred to the human and the robot releases its gripper accordingly. \\
\textbf{Heat Sandwich.} 
The ARX Lift-2 robot must open the oven with its left arm, pick up the plate containing the sandwich, place it into the oven using its right arm, and then close the oven with its left arm. 
This task assesses the model’s ability to operate articulated objects.
A trial is considered successful only if the plate is correctly inserted into the oven and the oven door is fully closed.
\\
\textbf{Sort Rubbish.}
The ARX Lift-2 robot must use its right arm to place all recyclable waste into the right bin and all non-recyclable waste into the left bin. 
This task evaluates the model’s ability to handle diverse object layouts and perform repetitive pick-and-place operations. 
A trial is considered successful only if all waste items are fully and correctly sorted. \\
\textbf{Sweep Trash.}
The ARX Lift-2 robot must grasp the dustpan with its right arm and the broom with its left arm. 
It then uses the broom to sweep all crumpled paper balls into the dustpan. 
Afterwards, the robot empties the dustpan into the left rubbish bin. 
Finally, it releases both grippers and returns to the home position. 
A trial is considered successful only if every step is finished successfully.
\\
\textbf{Sort Parts.}
The ARX Lift-2 robot must sort four types of small industrial components into four designated containers. 
These components include small nuts, assembly parts, and small screws. 
Each arm is responsible for sorting two categories. 
A trial is considered successful only if all components are placed into their correct containers. \\
\textbf{Unscrew Cap}
The ARX AC One robot must grasp the tea bottle with its left arm and move it to the designated middle zone. 
It then uses its right arm to approach the bottle cap and unscrew it. 
A trial is considered successful only if the cap is fully removed. \\
\textbf{Fold Cloths.}
The ARX AC One robot must fold the cloth into its designated final shape with both hands. 
A trial is considered successful only if the cloth is folded correctly.
\textbf{Zip Bag.}
The ARX AC One robot must use its left arm to open the bag, place all designated objects inside, and then zip it up.
A trial is considered successful only if the bag is fully and correctly zipped. \\

\subsection{Sim-to-real Task Description}
\textbf{Flip Package.}
A package is placed on the conveyor and moves toward the robot.
The ARX Lift-2 robot must grasp the package with its right arm, flip it over, and place it back onto the conveyor.
It must then grasp the package with its left arm and scan the QR code using the robot-mounted camera.
A trial is considered successful only if all steps are completed correctly.\\
\textbf{Instructional Pick.}
Eight types of objects are placed on the table.
A trial is considered successful only if the robot correctly picks the target object specified by the command.
\textbf{Sort Rubbish.}
This is the same task as described before. \\
\textbf{Wipe Stain.}
The ARX Lift-2 robot uses its left arm to pick up the towel and wipe stains located in one or two clusters.
A trial is considered successful only if all stains are completely removed. \\
\textbf{Sandwich.} 
The ARX Lift-2 robot uses its right arm to grasp a piece of bread and place it on the plate. 
It then uses its left arm to grasp a piece of beef and place it on the bread, followed by using the right arm again to place another piece of bread on top of the beef.
A trial is considered successful only if the sandwich is assembled correctly and neatly. \\
\textbf{Box.} 
The ARX Lift-2 robot sequentially closes the box lids with its right and left arms. 
A trial is considered successful upon complete closure.\\
\textbf{Microwave.}
The ARX Lift-2 robot uses the right arm to close the microwave lid.
A trial is considered successful upon complete closure. \\
\textbf{Pack.}
The ARX Lift-2 robot manipulates objects and places them into a box using its right and left arms. 
A trial is considered successful only after all objects have been placed inside.\\
\textbf{Sweep.}
This is the same task as described before.\\
\textbf{Handover.} 
The ARX Lift-2 robot uses its left arm to pick up a long-shaped object and hands it over to the right arm, which then places it into the box. 
A trial is considered successful only upon the object's transfer into the box.

\onecolumn
\begin{lstlisting}[style=yamlstyle, caption={\textbf{A Task Config Example on \textit{Sort Tray On Rack}.}}]
defaults:
  - _self_
  - world
  - logger
  - ../arenas@arena: scene_arena
  - ../cameras@astra: astra
  - ../cameras@realsense_d455_v3: realsense_d455_v3

name: banana_base_task
asset_root: assets
task: BananaBaseTask
task_id: 0

offset: null
render: True

env_map:
  envmap_lib: envmap_lib
  apply_randomization: True
  intensity_range: [4000, 7000]
  rotation_range: [0, 180]

robots:
  -
    name: "split_aloha"
    target_class: SplitAloha
    path: "split_aloha_mid_360/robot_task13.usd"
    euler: [0.0, 0.0, 90.0]
    robot_file:
      - curobo/src/curobo/content/configs/robot/piper100_left_arm.yml
      - curobo/src/curobo/content/configs/robot/piper100_right_arm.yml
    left_joint_home: [0.00484993, 0.34198609, -0.14007858, 0.01680429, 0.14391101, -0.00252178]
    right_joint_home: [0.00484993, 0.34198609, -0.14007858, 0.01680429, 0.14391101, -0.00252178]
    left_joint_home_std: [0.12513939, 0.24539099, 0.24468172, 0.23398885, 0.2710117, 0.21726329]
    right_joint_home_std: [0.12513939, 0.24539099, 0.24468172, 0.23398885, 0.2710117, 0.21726329]

objects:
  -
    name: arcode_plate_blue
    path: assets/plate/plate_blue/Aligned_obj.usd
    target_class: RigidObject
    dataset: arcode
    category: plate
    prim_path_child: Aligned
    translation: [0.0, 0.0, 0.0]
    euler: [90.0, 0.0, 0.0]
    scale: [1.0, 1.0, 1.0]
    apply_randomization: True
  -
    name: arcode_plate_shelf
    path: assets/plate_shelf/shelf_0/Aligned_obj.usd
    target_class: RigidObject
    dataset: arcode
    category: plate
    prim_path_child: Aligned
    translation: [0.0, 0.0, 0.0]
    euler: [90.0, 0.0, 0.0]
    scale: [1.0, 1.0, 1.0]
    apply_randomization: False

regions:
  -
    object: ${robots.0.name}
    target: table
    random_type: A_on_B_region_sampler
    random_config:
      pos_range: [
        [0.0, -0.86, -0.765], 
        [0.0, -0.86, -0.765]
      ]
      yaw_rotation: [0.0, 0.0]
  -
    object: arcode_plate_blue
    target: table
    random_type: A_on_B_region_sampler
    random_config:
      pos_range: [
        [0.125, -0.20, 0.005], 
        [0.25, -0.10, 0.005]
      ]
      yaw_rotation: [0, 0]
  -
    object: arcode_plate_shelf
    target: table
    random_type: A_on_B_region_sampler
    random_config:
      pos_range: [
        [-0.25, -0.20, 0.005], 
        [-0.15, -0.10, 0.005]
      ]
      yaw_rotation: [0, 0]

cameras:
  -
    name: ${robots.0.name}_hand_left
    translation: [0.0, 0.08, 0.05]
    orientation: [0.0, 0.0, 0.965, 0.259]
    camera_axes: usd
    params: ${astra}
    parent: "${robots.0.name}/split_aloha_mid_360_with_piper/fl/link6"
    apply_randomization: True
    max_translation_noise: 0.03
    max_orientation_noise: 5.0

  -
    name: ${robots.0.name}_hand_right
    translation: [0.0, 0.08, 0.04]
    orientation: [0.0, 0.0, 0.972, 0.233]
    camera_axes: usd
    params: ${astra}
    parent: "${robots.0.name}/split_aloha_mid_360_with_piper/fr/link6"
    apply_randomization: True
    max_translation_noise: 0.03
    max_orientation_noise: 5.0

  -
    name: ${robots.0.name}_head
    translation: [0.0, -0.00818, 0.1]
    orientation: [0.658, 0.259, -0.282, -0.648]
    camera_axes: usd
    params: ${realsense_d455_v3}
    parent: "${robots.0.name}/split_aloha_mid_360_with_piper/top_camera_link"
    apply_randomization: True
    max_translation_noise: 0.03
    max_orientation_noise: 5.0

data:
  save_root_path: "InternData-A1/sim/raw_data"
  task_dir: "Sort Tray On Rack"
  language_instruction: "Pick the plate, make the handover and place it on the water cooling holder"
  detailed_language_instruction: "Pick the plate with the right arm, make the handover to the left arm, and then place it on the water cooling holder."
  collect_info: ""
  version: "v3.0, head camerea 1280x720, wrist 640x480, y 45 degrees"
  update: True
  max_episode_length: 4000

skills:
  -
    split_aloha:
      -
        right:
          -
            name: pick
            objects: [arcode_plate_blue]
            filter_x_dir: ["upward", 90, 45]
            filter_y_dir: ["forward", 40]
            filter_z_dir: ["downward", 110, 140]
            t_eps: 0.01
            o_eps: 1
            close_wait_steps: 10
            post_grasp_offset_min: 0.1    
            post_grasp_offset_max: 0.1  
            direction_to_obj: right

          -
            name: goto__pose
            frame: robot
            gripper_action: close_gripper
            translation: [0.3, 0.13, 0.15]
            quaternion: [-0.15, -0.37, -0.84, -0.36]

      -
        left:
          -
            name: pick
            objects: [arcode_plate_blue]
            filter_y_dir: ["upward", 40]
            filter_z_dir: ["forward", 90, 45]
            close_wait_steps: 10
            t_eps: 0.01
            o_eps: 1
            post_grasp_offset_min: 0.0 
            post_grasp_offset_max: 0.0    
            direction_to_obj: left

      -
        left:
          - name: gripper__action
            action_type: close
        right:
          - name: gripper__action
            action_type: open

      -
        right:
          - name: home

      -
        left:
          -
            name: place
            place_direction: vertical
            objects: [arcode_plate_blue, arcode_plate_shelf]
            filter_y_dir: ["upward", 60, 0]
            filter_z_dir: ["forward", 90, 30] 
            position_constraint: object
            x_ratio_range: [0.5, 0.5]
            y_ratio_range: [0.8, 0.8]  
            align_pick_obj_axis: [0, 1, 0]
            align_place_obj_axis: [0, 0, 1] 
            align_obj_tol: 10 
            pre_place_z_offset: 0.15
            place_z_offset: 0.01
\end{lstlisting}
\label{tab:supp:task_config}
\twocolumn

\end{document}